# Advanced Knowledge Extraction of Physical Design Drawings, Translation and conversion to CAD formats using Deep Learning

Jesher Joshua M[1], Ragav V[2], Syed Ibrahim S P [C*3]

*Vellore Institute of Technology, Chennai Campus, Chennai – 600 127, India*

**Abstract**

The maintenance, archiving and usage of the design drawings is cumbersome in physical form in different industries for longer period. It is hard to extract information by simple scanning of drawing sheets. Converting them to their digital formats such as Computer-Aided Design (CAD), with needed knowledge extraction can solve this problem. The conversion of these machine drawings to its digital form is a crucial challenge which requires advanced techniques. This research proposes an innovative methodology utilizing Deep Learning methods. The approach employs object detection model, such as Yolov7, Faster R-CNN, to detect physical drawing objects present in the images followed by, edge detection algorithms such as canny filter to extract and refine the identified lines from the drawing region and curve detection techniques to detect circle. Also ornaments (complex shapes) within the drawings are extracted. To ensure comprehensive conversion, an Optical Character Recognition (OCR) tool is integrated to identify and extract the text elements from the drawings. The extracted data which includes the lines, shapes and text is consolidated and stored in a structured comma separated values(.csv) file format. The accuracy and the efficiency of conversion is evaluated. Through this, conversion can be automated to help organizations enhance their productivity, facilitate seamless collaborations and preserve valuable design information in a digital format easily accessible. Overall, this study contributes to the advancement of CAD conversions, providing accurate results from the translating process. Future research can focus on handling diverse drawing types, enhanced accuracy in shape and line detection and extraction.

*Keywords: CAD conversion; deep learning; design drawings; digital formats; optical character; knowledge extraction;*

## 1. Introduction

Engineering and design drawings are prevalently used in various continuous process industries such as power, cement, oil and gas etc., to depict their structural and dimensional technical schematic configuration, process flow, circuits, device functions etc. Preserving them from paper forms to digitized media is important for retrieving the needed information retrievable easily in needed format as they are rich source of technical configuration information. Decoding the scanned drawings typically needs set of image processing techniques to detect and classify various symbols and components.

Detecting different components in automated method, may reduce major portion of effort and time. The automated design extraction, validation and storage in standard format enables users to retrieve the count, connections, location, accuracy of components. The proposed system intakes target design images as input and detects various components, objects and text present in the image and storing the extracted objects and texts in CAD file format. Though there are recent advances in image processing and computer vision techniques, the end-to-end solution from transforming the physical diagrams information to structured formats in still incomplete to develop into a matured solution with needed accuracy.

This paper proposes a comprehensive generic framework for digitizing engineering drawings and identify component in the drawings by various algorithms and deep learning techniques to detect edges, lines, curves, objects of different shapes and texts and the extracted information is systematically stored in the needed format.

[1.] *Under Graduate Student, School of Computer Science and Engineering, E-mail address:* jesherjoshua.m2021@vitstudent.ac.in
[2] *Under Graduate Student, School of Computer Science and Engineering, E-mail address:* ragav.v2021@vitstudent.ac.in
[3*] *Corresponding author.*
*Professor, School of Computer Science and Engineering, E-mail address:* syedibrahim.sp@vit.ac.in

## 2. Analysis of existing models

Digitizing design drawings requires effective image processing techniques and pass through sequence of preprocessing, objects detection, classification, text extraction and contextualizing the relations of the objects represented. We find literatures more than thirty-five years ago, like Ishii et al.[1] proposing solutions to digitize piping and instrumentation systems drawing digitization. Large amount of research work, findings and proposed solutions exists which are mostly domain specific or applicable for specific needs. For example few focus on definite components of the digitization process, like symbols identification [2-3] & component classification [4], geometric labels extraction such as length, thickness, radius etc.[5] While others elaborate optical character recognition (OCR) techniques for text extraction [6-7], and on direct digitization of drawings to CAD files[8-9]. More complex and real time engineering drawing utilized in actual industries are attempted to be digitized and component extracted with stronger contextulization models proposed [10]. Most advanced models like, Gellaboina et al. put forth a symbol detection method with learning strategy based on the recurrent training of a ANN using the Hopfield model.[12]

## 3. Proposed Solution

A custom dataset of 200 representative design diagrams images were collected, cleansed, accurately annotated, trained and tested using the proposed model. The proposed solution architecture consists of two sections, the feature extraction and conversion of the information. Post training and stabilization the model, it can be used to detect the drawing objects of any scanned image (jpeg, png, pdf etc.) sent as the input to be processed and get the CAD file format output. For each input image, detection of the drawing region in the image, applying filters and preprocessing the image for further collection of information about the edges, lines, curves and the ornaments present in the drawing region are performed. Subsequently, the text legends corresponding to the drawing present in the image is also identified using the Optical Character Recognition (OCR) tool and the consolidated coordinates of these objects are stored in a comma separated values (.csv) format and further converted into a Scalable Vector Graphics (SVG) or a CAD file format.

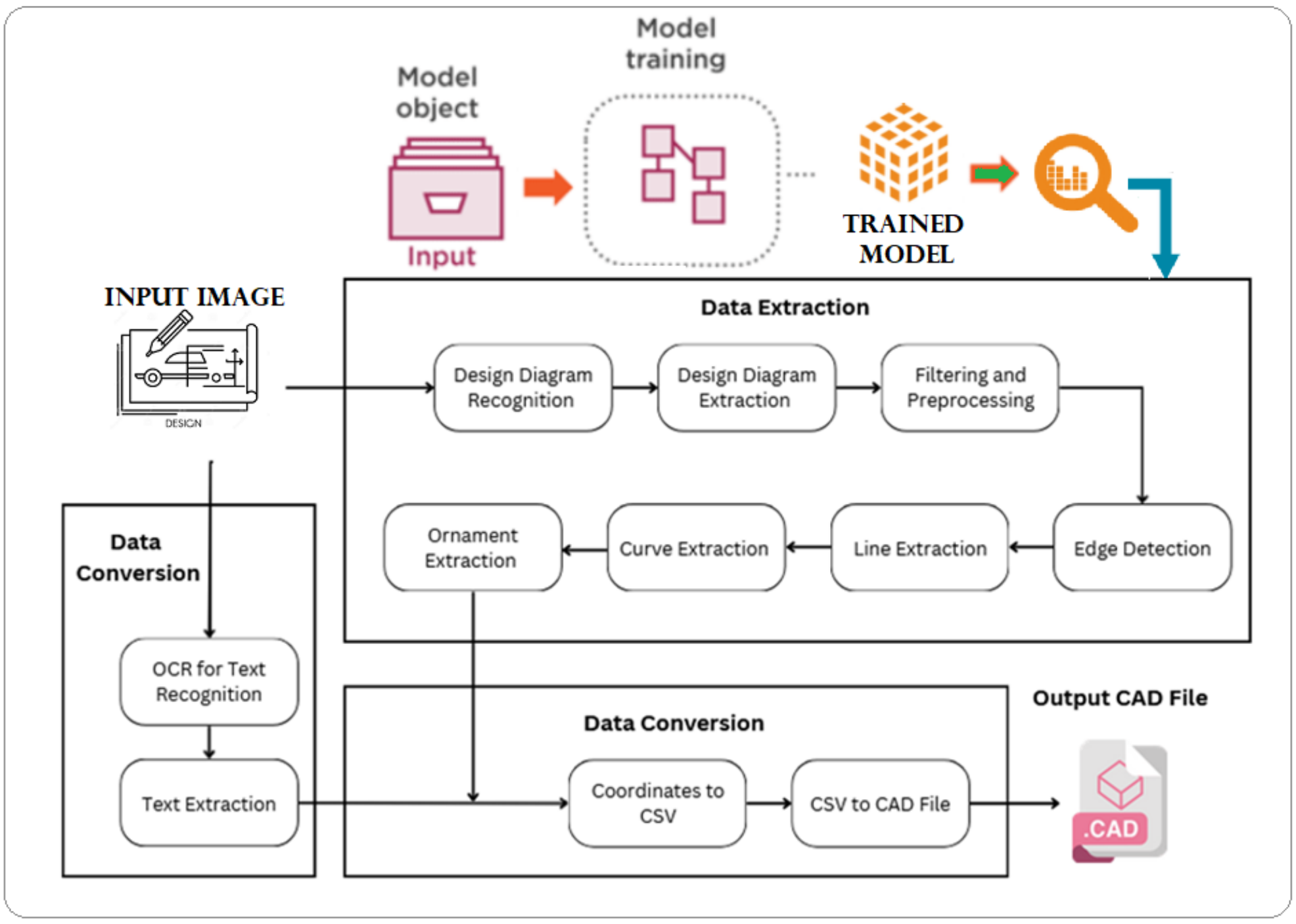

Fig. 1. Proposed solution Architecture

Overall, The Image of the physical design diagram is passed as the input to the proposed model which passes through the data extraction section, collects all the information, coordinates of the objects within the design and in the data conversion section, where the text data is obtained and stored as structured format along with the information collected and finally gets converted into a digitalized format.

Summarizing the above sections, the below algorithm for digital drawing data extraction and conversions to CAD formats depicts the sequence of activities performed.

**Algorithm 1 : Digital Drawing Conversion**

**Input:** Path to PDF file;
**Output:** Extracted digital information from the drawing;

DigitalDrawingConversionPDFFilePath CREATEDATASET() ;
`// Extract images from the PDF document using the FITZ library`

READ and preprocess the image files using the cv2 and PIL libraries;
PERFORMEDGEDETECTION() ; `// Detect edges in the image using the Canny edge detection`
GETLINES() ; `// Detect lines in the image using the Hough Transform`
GETCIRCLES() ; `// Detect circles in the image using the Hough Transform`
YOLOLIGHTS() ; `// Recognize decor (lights) in the image using the YOLO model for lights`
YOLODIMLINES() ; `// Recognize dimension lines in the image using the YOLO model for dimension lines`
TXTTOCSV(*mode* = "lights") ; `// Convert YOLO output for decor to CSV format`
TXTTOCSV(*mode* = "dimlines") ; `// Convert YOLO output for dimension lines to CSV format`
PERFORMOCR() ; `// Perform Optical Character Recognition (OCR) to extract text from the image`
EXTRACTTEXT() ; `// Extract relevant information from the OCR output`

*3.1 Data extraction*

*3.1.1 Design diagram recognition*

The fundamental step of detecting whether a meaningful drawing is present in the input image for the model to move forward with the remaining sections, is performed using custom trained Deep learning models such as the YOLOv6, YOLOv7, YOLOv8 and its results were evaluated and that of YOLOv7 and YOLOv8 were compared, of which YOLOv7 produced satisfactory results in detecting the drawing region by drawing bounding boxes.

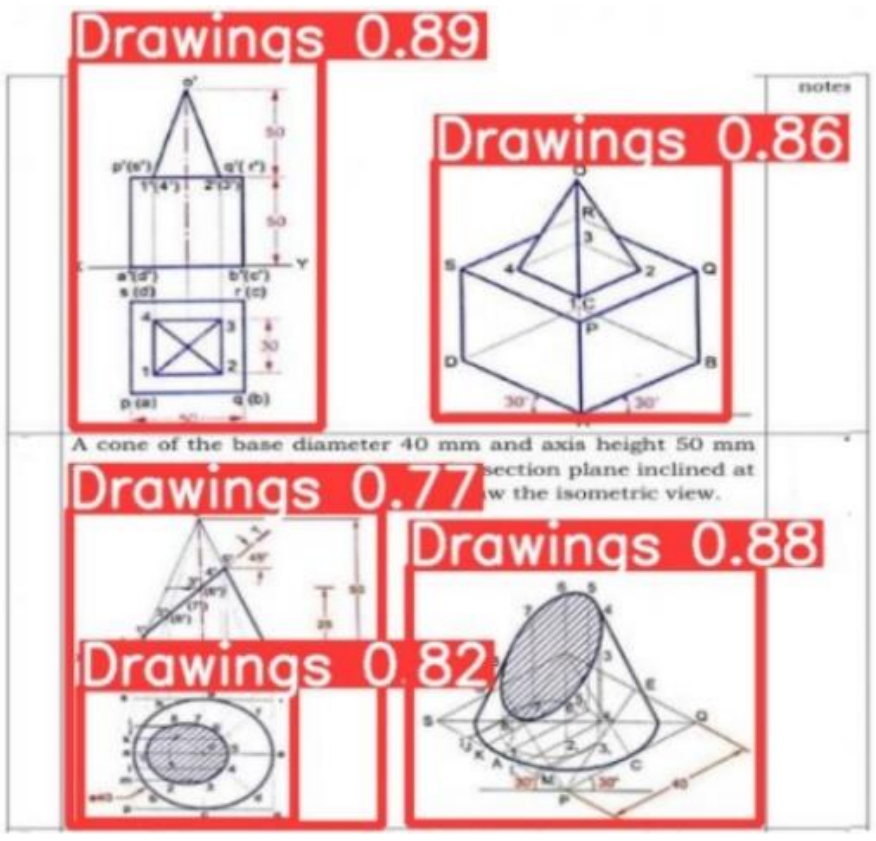

Fig. 2 Design Diagram Recognition

From below figures 3 to 6, we find that the Yolo7 is more suitable for recognizing the design diagrams as compared to other models 6 & 8. The Accuracy, F1 score, Precision and recall results are encouraging and best suited to produce accurate prediction of diagram identification.

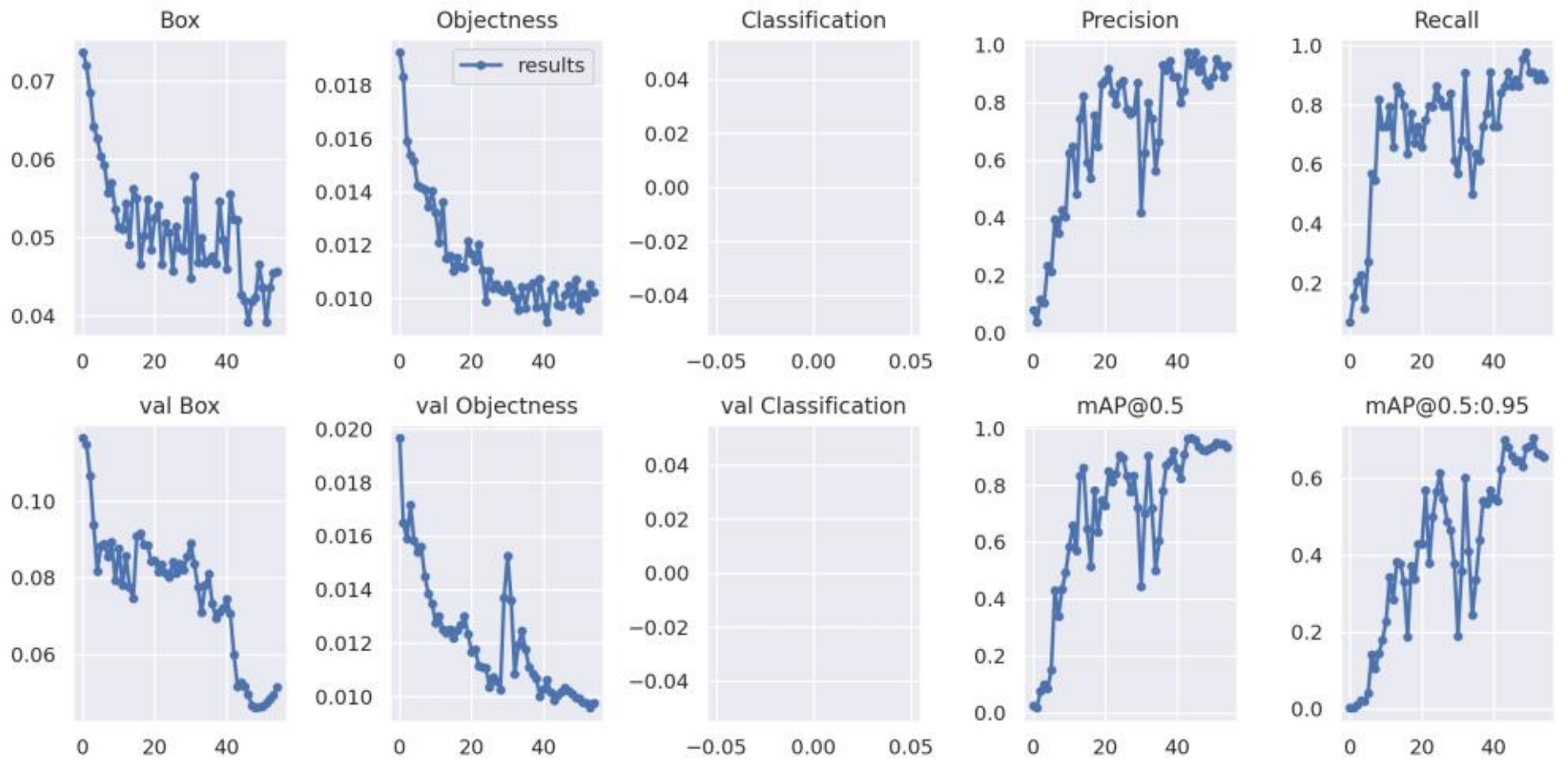

Fig. 3. Yolo7 Accuracy results

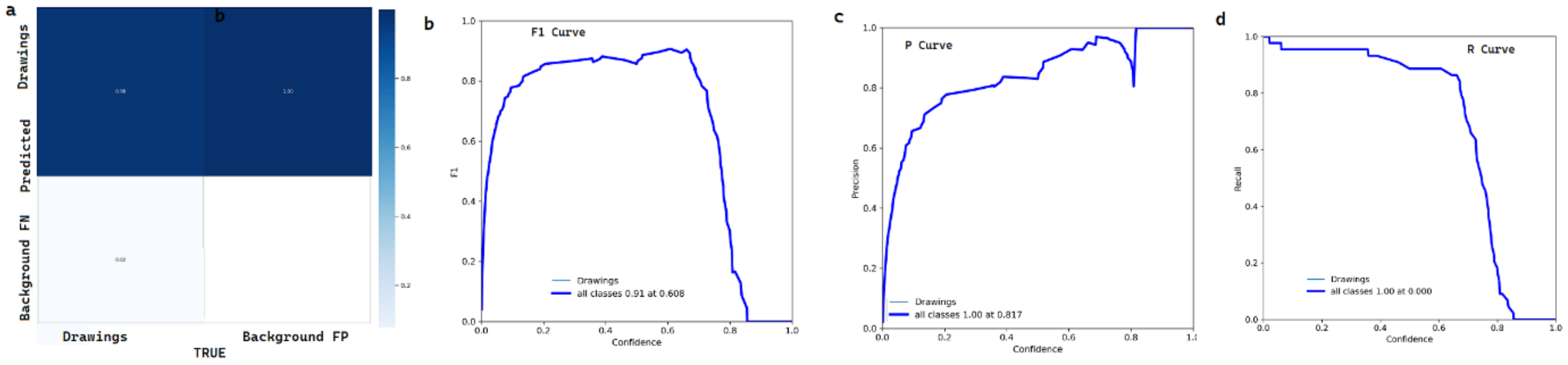

Fig. 4. Yolov7 : (a) Confusion Matrix, (b) F1 Curve, (c) Precision Curve, (d) Recall Curve

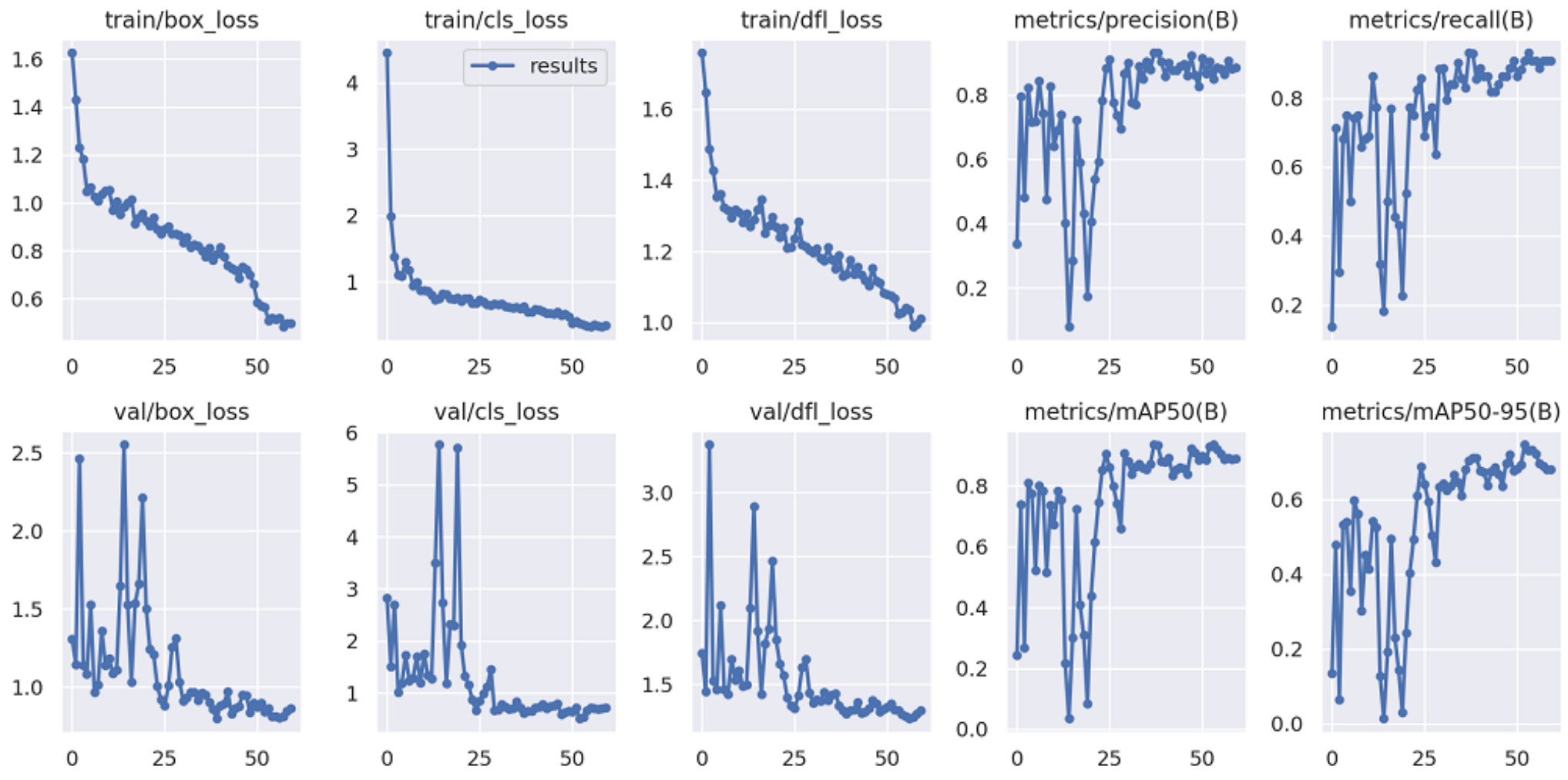

Fig. 5. Yolov8 Accuracy results

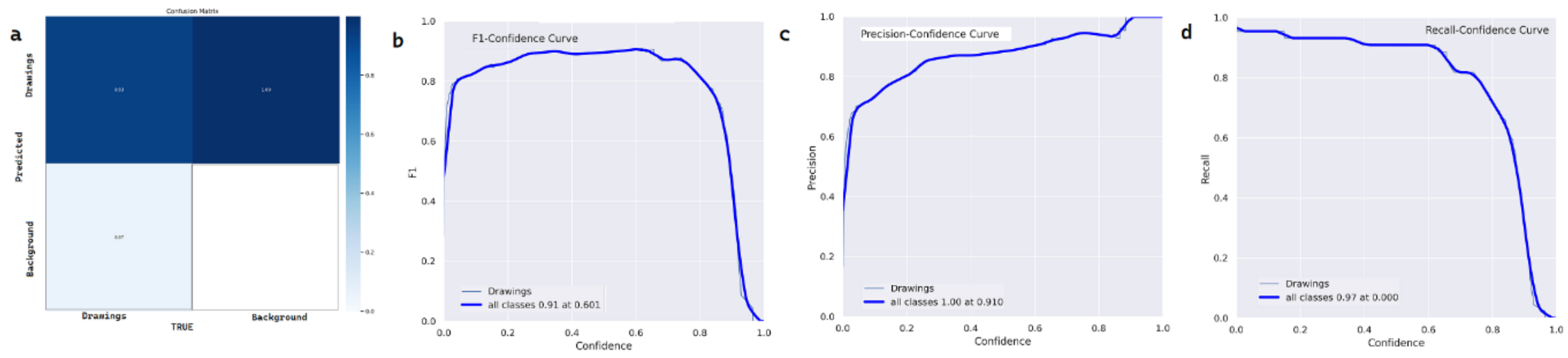

Fig. 6. Yolov8 : (a) Confusion Matrix, (b) F1 Curve, (c) Precision Curve, (d) Recall Curve

*3.1.2 Design diagram extraction*

The Region of Interest (ROI) where the drawing is present in the input image is precisely derived and the coordinates of the recognized ROI are marked from the bounding boxes to apply the further steps of preprocessing and data extraction.

*3.1.3 Filtering and preprocessing*

On the drawing region extracted, the Canny filter is applied for edge detection, magnification and the required contrast enhancements, image alignment along with noise reduction are performed as a preprocessing step using OpenCV. With filtering helps the symbols, shapes and characters in the dataset, creating a better and enhanced version of the image, even if the input images are of poor image quality and occulated. Canny filtering has 4 steps like noise Reduction, gradients calculation, No maximum suppression and Thresholding with hysteresis.

*3.1.4 Edge Detection, Line and Curve Extraction*

Zhang et al.[13] proposed 2 types of structural feature extraction viz. contour and region based. Wenyin et al.[[14] presented a structural feature extraction method based on analysing all possible pairs of line segments composing the symbol.

Feature extraction technique which operates in a parameter space can be used to detect objects like Line or Circle. Each point in the space represents a possible line or circle which are detected with below 5 transformation steps:

Edge Detection: Before applying the Transform, an edge detection algorithm, such as the Canny edge detector, is applied to the image. This step identifies the edges of objects in the drawing, used for shape detection.

Representation of Parameter Space: The lines in the image using a parameter space. For line detection, each pixel in the edge image denotes a possible point on a line. In this space, lines are denoted with the polar coordinate system. Each line in the image maps to a sinusoidal curve in the space.

Accumulation: For each edge pixel in the edge image, the Transform algorithm finds the parameters ($\rho$ and $\theta$) of all possible lines that pass through that point. This related accumulator cells are incremented in the space. The accumulator cells record the count of votes for each line parameter.

Peaks finding: Post Accumulation, the algorithm finds the peaks in accumulator cells. These peaks denote the lines that have received the most votes, indicating the most possible lines in the image. The threshold for finalizing a peak which can be adjusted to control the sensitivity of detection of line.

Extraction of Lines: The detected peaks in the space are used to extract lines. Line parameters converted back to Cartesian coordinates in the original image space.

The is a useful technique for detecting lines, but is computationally intensive. It is robust against noise and partial occlusions, making it suitable for engineering drawing analysis.

The Key functions for line detection used are:

- put_box(yolo_bbox) - To convert bounding box coordinates from the YOLO format to the format required for further processing.
- create_dataset(fpath) : To extract images from PDF files and saves them in the input folder, using a library like fitz.
- get_lines(edges, img): To detect lines in the edge image and draws them on the original image.
- get_circles(edges, img): To detect circles in the edge image using the HoughCircles algorithm.
- yolo_lights(image_path): YOLO model trained specifically for decor (light) recognition in engineering drawings.
- yolo_dimlines(image_path): To recognize dimension line in engineering drawings.

With the above steps the lines present on the design are extracted from the filtered image which contains the detected and enhanced edges using custom trained YOLOv which draws the lines present from the edges. Along with the lines, the dimension lines corresponding to the breadth and width of the design are also identified.

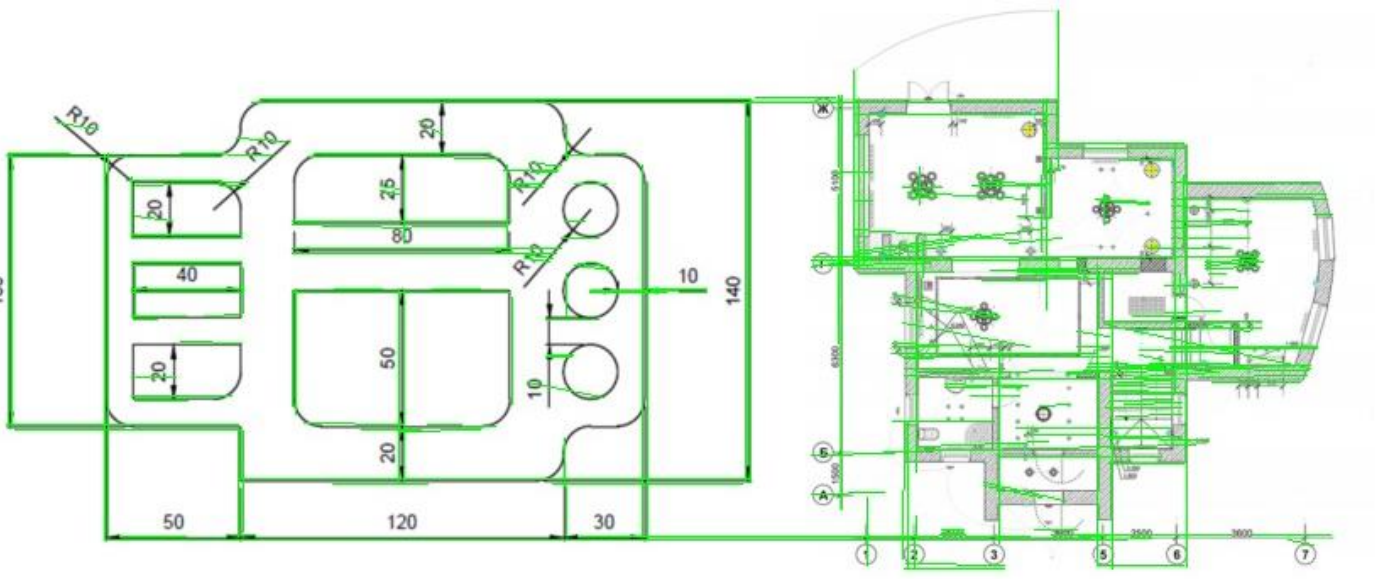

Fig. 7. Line extraction

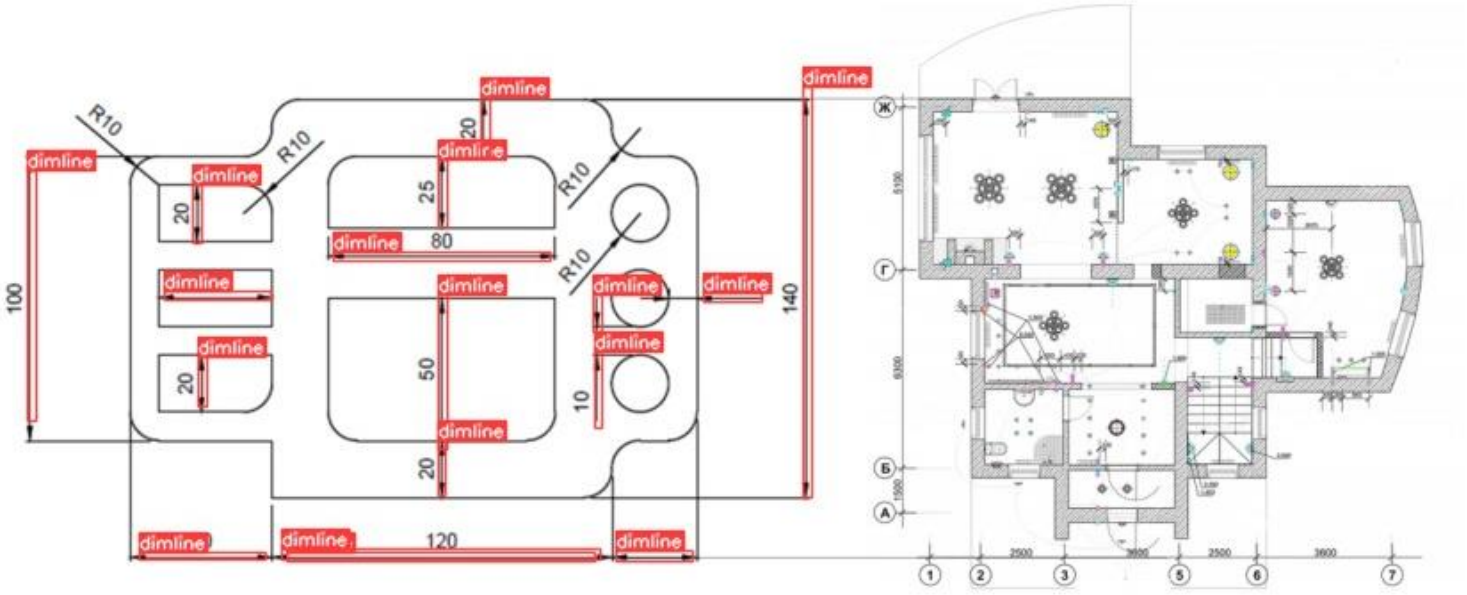

Fig. 8. Dimension line detection

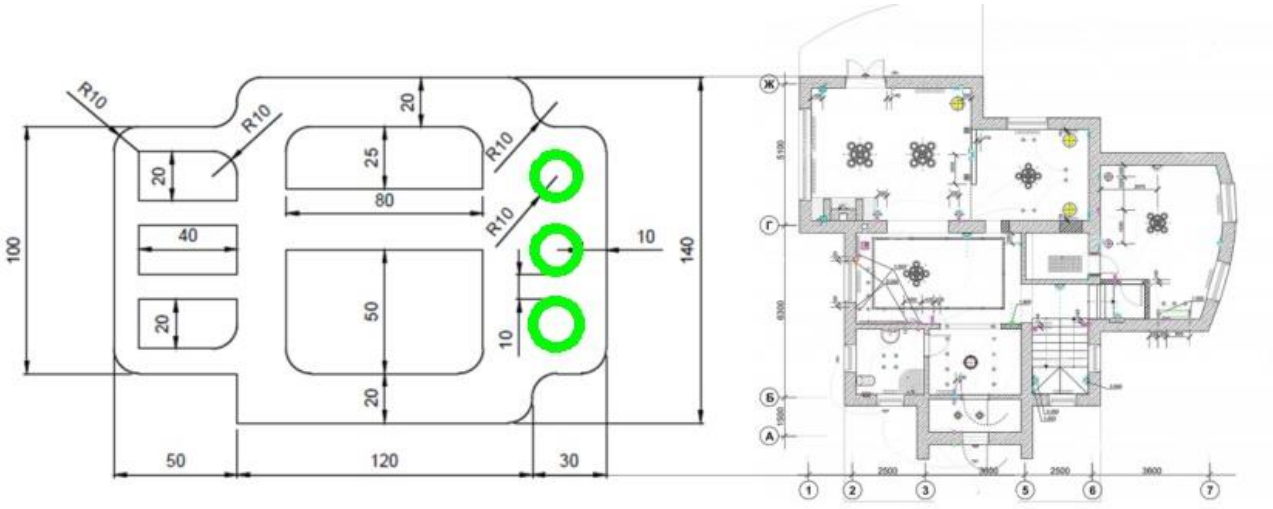

Fig. 9. Resulting extraction of the curves (circles) from the ROI using the YOLO

### *3.1.5 Ornament Extraction*

Ornaments are complex shapes present in the drawing which are again extracted using Yolo model.

Common ornaments that were annotated and detected using the YOLO model were,

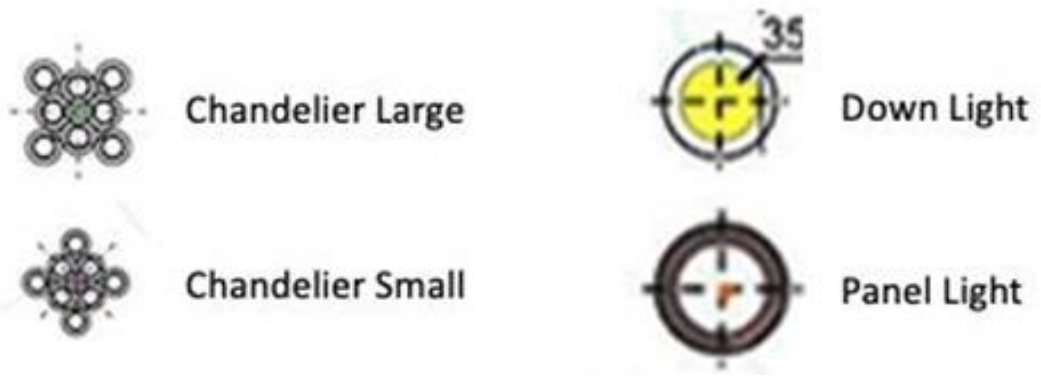

Fig. 10. Ornaments annotated in the custom dataset.

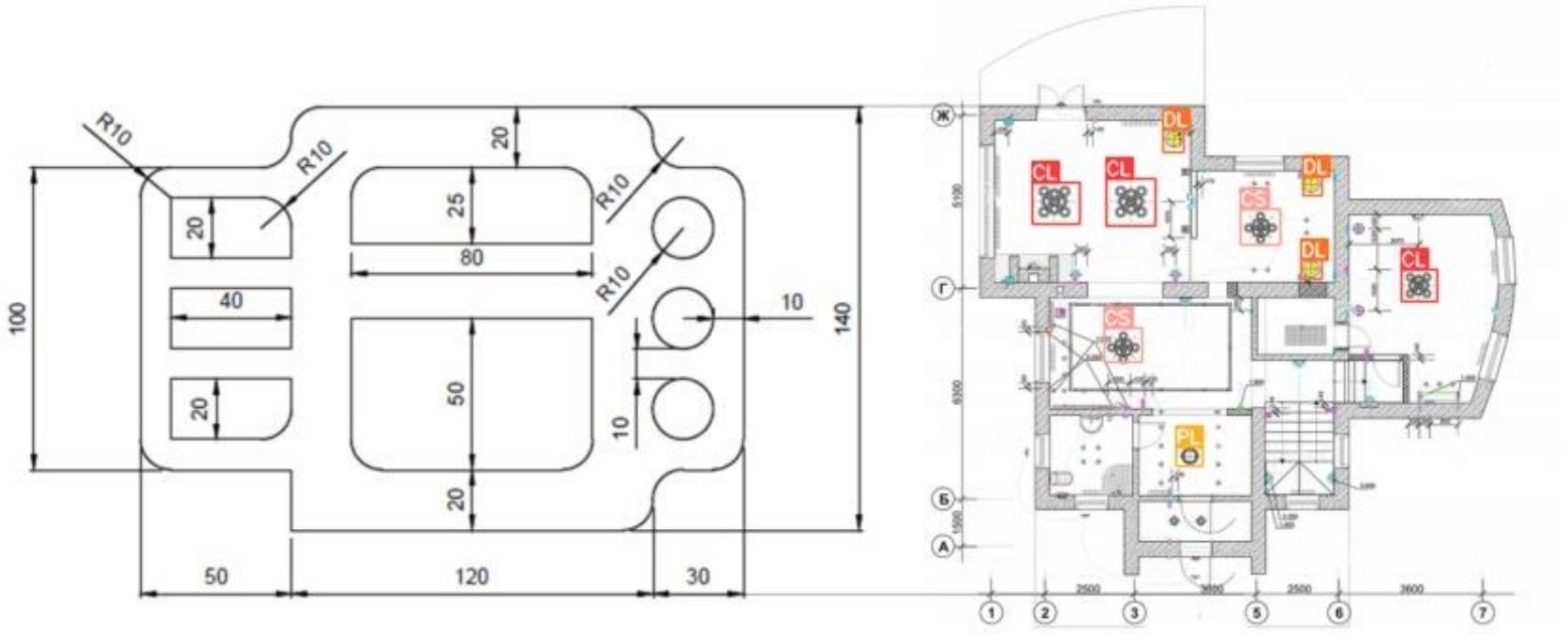

Fig. 11. Resulting extraction of the ornaments from the ROI using the YOLOv model

*3.2 Data conversion*

The dimensions of the lines, curves, shapes and ornaments were collected from the model and stored with their coordinates in this section of data extraction. These data along with addition information should be stored in a structured format for further use which is performed on the data conversion section where the texts corresponding to the drawing are extracted to add comprehensive meaning to the design digitally converted and the consolidated data is stored in a .csv file format and further converted to a CAD format.

*3.2.1 OCR for text recognition*

For text interpretation from drawings, there are reviews on OCR for printed documents[15,16] and multiple open source OCR software available such as Tesseract6 and PhotoOCR[17] being widely used. The design diagrams typically contain textual information such as legends, descriptions, dimensional details of the components of the design. Primarily, the text objects are identified and bounding boxes over the objects are drawn to facilitate cropping and further the text is extracted from the cropped image using the Optical Character Recognition (OCR).

a

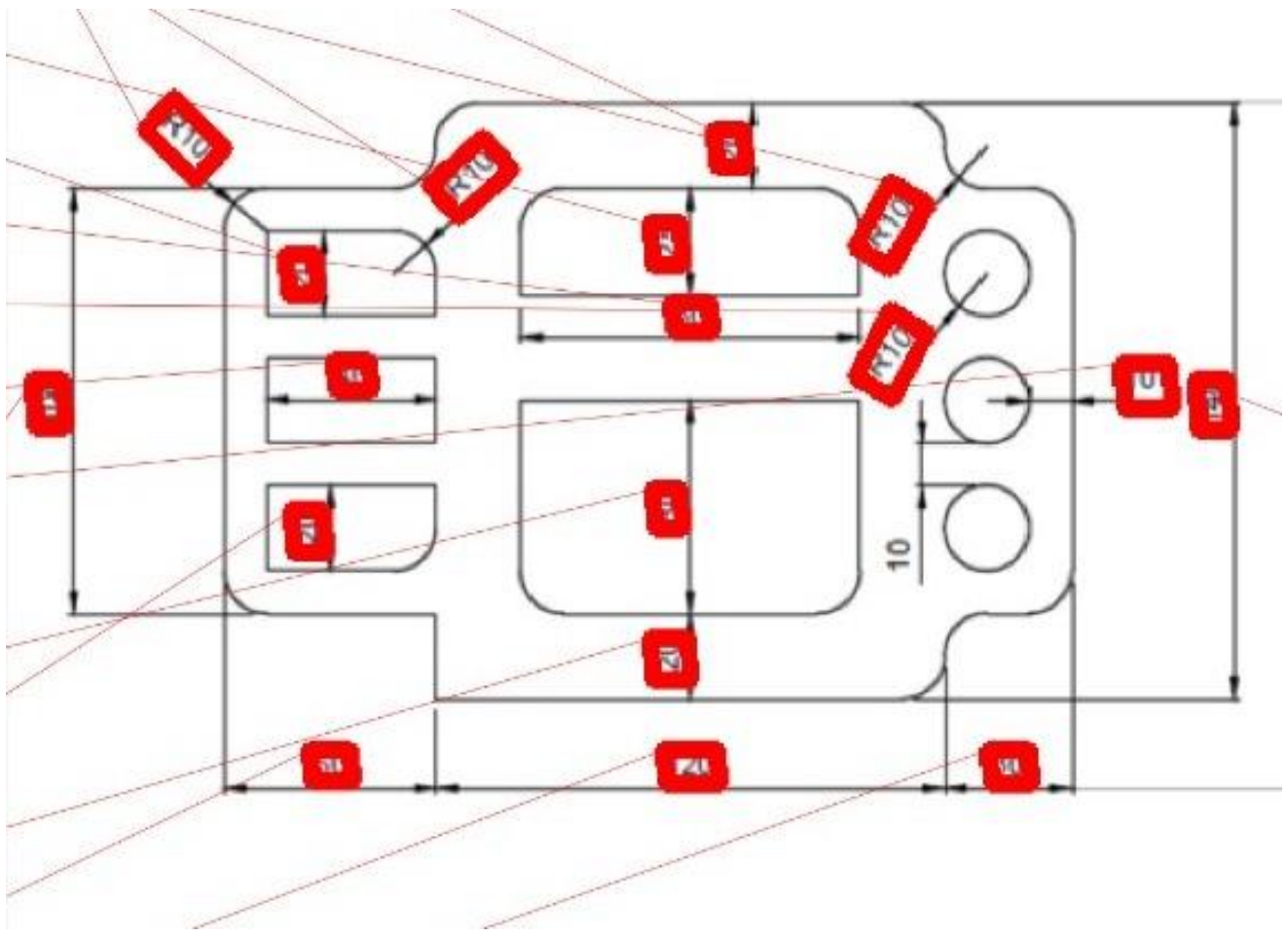

b

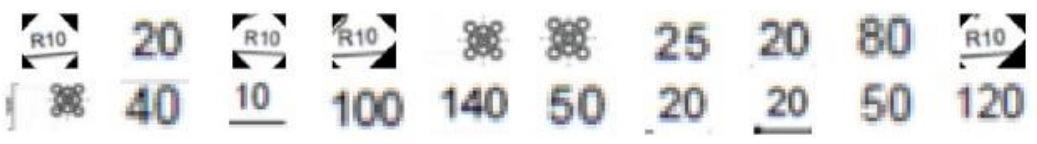

Fig. 12. (a) Bounding Boxes over the recognized text objects; (b) Images of the text objects cropped from the ROI.

### *3.2.2 Consolidated conversion of data extracted*

From the above models, the various shapes detected are extracted, consolidated and converted to csv file format for each line, circle, lights etc., with coordinates of individual items saved.

**Converted csv File (Circle coordinates)**

| | x | y | radius |
|---|---|---|---|
| **Circle** | 368 | 280 | 14 |
| **Circle** | 368 | 232 | 14 |
| **Circle** | 368 | 328 | 15 |

**Converted csv File (Dimension line coordinates)**

| | x1 | y1 | x2 | y2 |
|---|---|---|---|---|
| **Dimension Line** | 342 | 278 | 347 | 299 |
| **Dimension Line** | 170 | 423 | 343 | 428 |
| **Dimension Line** | 404 | 277 | 437 | 282 |
| **Dimension Line** | 255 | 278 | 260 | 365 |
| **Dimension Line** | 355 | 422 | 398 | 428 |

**Extracted csv File (Lights Coordinates)**

| | x1 | y1 | x2 | y2 |
|---|---|---|---|---|
| **PL** | 629 | 348 | 643 | 358 |
| **CS** | 591 | 285 | 611 | 303 |
| **CS** | 663 | 222 | 684 | 241 |
| **CL** | 748 | 254 | 767 | 271 |
| **DL** | 696 | 205 | 706 | 215 |

**Extracted csv File (Lines Coordinates)**

| | x1 | y1 | x2 | y2 |
|---|---|---|---|---|
| **Line** | 81 | 427 | 686 | 427 |
| **Line** | 361 | 425 | 574 | 425 |
| **Line** | 168 | 169 | 645 | 169 |
| **Line** | 173 | 167 | 463 | 167 |
| **Line** | 346 | 394 | 458 | 394 |
| **Line** | 24 | 201 | 150 | 201 |

The component objects coordinate details in the csv files can be converted to CAD/SVG format for standardized storage for easier information retrieval and processing.

## 4. Result Analysis

Below table lists down the number of drawing objects detected in single image and its accuracy in detection capability.

Table.1 Result Analysis

| S.No | Component | Ground Truth | Proposed Solution |
|---|---|---|---|
| 1. | No. of Circles | 3 | 3 |
| 2. | No. of Ornaments | 8 | 8 |
| 3. | No. of Dimension Lines | 20 | 16 |
| 4. | No. of Text Regions | 18 | 17 |
| 5. | Accuracy of Text | 100% | 93% |

This suggests that the proposed model has provided satisfactory results on detecting and extracting the prescribed objects from the input image.

Below is the resultant converted CAD file of the extracted and converted file with output shapes detected and coordinates captured with all object types consolidated for easier knowledge extraction and retrival.

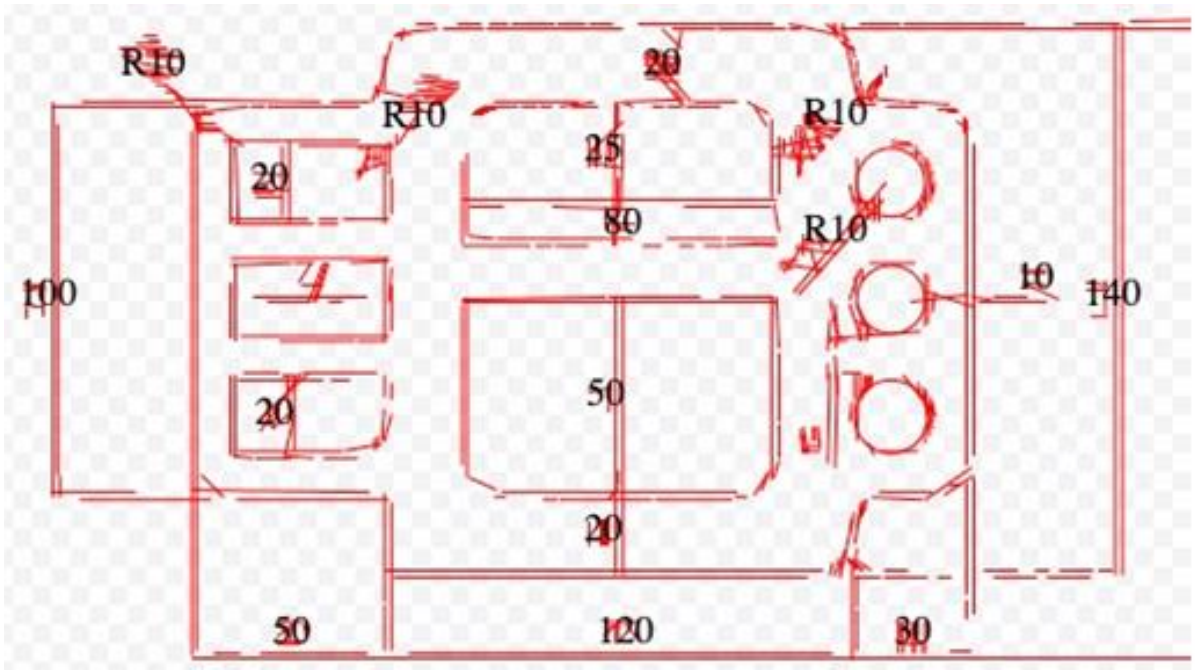

Fig. 13 Drawing attributes conversion to CAD format

**5. Conclusion**

The proposed design drawing knowledge extraction and digitization framework provides a unique prospect for relevant industries to leverage digitization of large volumes of diagrams in informing their decision-making process and future needs.

This research paper presents an innovative approach for converting machine/engineering drawings into CAD digital formats. The proposed methodology involves a series of techniques including object detection for ROI extraction, canny filtering and Hough transforms for edge detection and line extraction, Hough circles for curve identification, object detection for dimension line detection, and OCR for text region extraction and recognition.

The experimental results demonstrate the effectiveness of the proposed method in accurately converting hand-drawn engineering drawings to CAD digital formats. The utilization of object detection algorithms, specifically YOLO, for dimension line extraction and ROI identification improves the precision and efficiency of the conversion process. By leveraging canny filtering, Hough transforms, and Hough circles, the approach successfully captures and represents the various geometric shapes and curves commonly found in engineering drawings.

The integration of OCR enables the extraction of textual information from the drawings, enhancing the completeness and fidelity of the converted CAD models. The methodology proves its capability to handle complex drawings with different line styles, orientations, and shapes.

**6. Future Enhancements**

While the presented research has made significant progress in converting machine/engineering drawings to CAD digital formats, there are several areas that offer potential for future enhancement:

**YOLO Model Optimization**: Fine-tuning the YOLO model for dimension line extraction and ROI identification specifically for machine/engineering drawings could further improve accuracy and efficiency[18]. Customizing the model by expanding the dataset, retraining with specialized annotations, and adjusting hyperparameters would contribute to achieving better results.

**Advanced Line and Curve Extraction Techniques**: Exploring advanced algorithms, such as deep learning-based methods or hybrid approaches, can enhance the extraction of lines and curves from engineering drawings. These techniques may address challenges related to complex geometries, noisy or incomplete lines, and improve robustness.

**Incorporating Graphical Context**: Integrating graphical context information, such as spatial relationships between components in the drawing, can enhance the overall accuracy and understanding of the CAD conversion process. Techniques like graph-based representations or semantic segmentation can capture contextual information, leading to more precise and meaningful CAD conversions[19].

**Real-Time Conversion and Interactive Editing**: Extending the methodology to support real-time conversion and interactive editing within CAD software would enhance usability and practicality. Developing efficient algorithms and optimizing the implementation for quick processing and seamless integration with existing CAD tools would be valuable for users.[20]

**Domain-Specific OCR Improvements**: Customizing OCR algorithms for technical text extraction can enhance accuracy and reliability. Training the OCR model on domain-specific datasets tailored to technical drawings or engineering texts would improve recognition of specialized font styles, abbreviations, and notations.

**User Interface and Visualization:** Designing an intuitive user interface that provides visual feedback during the conversion process, such as highlighting extracted components, identifying errors, and offering suggestions for manual corrections, would improve the quality and accuracy of CAD conversions.

In conclusion, this research paper introduces a novel approach for converting machine/engineering drawings to CAD digital formats. By combining object detection, canny filtering, Hough transforms, Hough circles, and OCR techniques, the proposed methodology demonstrates its capability to automate the CAD conversion process. Future enhancements focusing on YOLO model optimization, advanced line and curve extraction techniques, incorporating graphical context, real-time conversion and interactive editing, domain-specific OCR improvements, and user interface and visualization will further refine and extend the methodology's capabilities, making it an invaluable tool for automated CAD conversion.